\documentclass[conference,a4paper]{IEEEtran}

\makeatletter
\def\ps@IEEEtitlepagestyle{%
  \def\@oddfoot{\mycopyrightnotice}%
  \def\@evenfoot{}%
}
\def\mycopyrightnotice{%
  \parbox{\textwidth}{\centering\footnotesize
  Accepted for presentation at ICECCME 2026. Authors' accepted manuscript for arXiv posting;\\
  copyright may be transferred without notice after IEEE copyright processing.}%
  \gdef\mycopyrightnotice{}%
}

\IEEEoverridecommandlockouts
\usepackage{cite}
\usepackage{amsmath,amssymb,amsfonts}
\usepackage{graphicx}
\usepackage{textcomp}
\usepackage{xcolor}
\usepackage{booktabs}
\usepackage{array}
\usepackage{url}
\usepackage{eso-pic}
\usepackage{pgfplots}
\pgfplotsset{compat=1.18}
\usepgfplotslibrary{groupplots}
\usetikzlibrary{arrows.meta}
\newcommand{\figlabel}{\fontsize{7.2}{8.4}\selectfont}
\newcommand{\metriclabel}[2]{\begin{tabular}{@{}l@{}}#1\\#2\end{tabular}}

\def\BibTeX{{\rm B\kern-.05em{\sc i\kern-.025em b}\kern-.08em
    T\kern-.1667em\lower.7ex\hbox{E}\kern-.125emX}}

\newcommand\AtPageUpperMyright[1]{\AtPageUpperLeft{%
 \put(\LenToUnit{0.17\paperwidth},\LenToUnit{-2cm}){%
     \parbox{0.9\textwidth}{\raggedleft\fontsize{8}{11}\selectfont #1}}%
 }}%
\newcommand{\conf}[1]{%
\AddToShipoutPictureBG*{%
\AtPageUpperMyright{#1}
}
}

\begin{document}
\title{\vspace*{1cm} Self-Commitment Latency: A Reward-Free Probe for Prompted Implicit Hacking}

\author{\IEEEauthorblockN{Bonan Shen$^1$, Youting Wang$^2$, Dingyan Shang$^1$, Tao Ning$^3$}
\IEEEauthorblockA{\textit{$^1$Independent Researcher, $^2$Northeastern University, $^3$Syracuse University}\\
Emails: \{shenbonan2, ginkoin613, dingyanshang, ntgd1102\}@gmail.com}
}

\maketitle
\conf{\textit{Accepted for presentation at the International Conference on Electrical, Computer, Communications and Mechatronics Engineering (ICECCME 2026)\\
15-17 October 2026, Bali, Indonesia}}

\begin{abstract}
Implicit reward hacking is hard to audit when a language model's chain of thought appears benign: a final answer may be anchored by a prompt shortcut while the written reasoning still resembles ordinary problem solving. Verifier-based probes expose such behavior by measuring how early truncated reasoning contexts obtain high reward, but require a task-specific reward signal. This paper proposes a weaker-input alternative, \emph{self-commitment latency}, which measures how early a prompted reasoning context commits to the model's own final answer. We evaluate the probe in a controlled paired GSM8K setting using Qwen2.5-3B-Instruct-4bit, comparing ordinary prompts with prompts that include an answer hint. Hinted contexts commit substantially earlier and with lower uncertainty than honest contexts. The primary latency metric, first-commitment latency at threshold 0.8, reaches AUROC 0.878; supporting whole-curve summaries reach AUROC 0.926 for commitment range and 0.904 for mean uncommitted mass. The signal is stronger when both prompt conditions answer correctly and remains stable across thresholds. These results show that shortcut-available reasoning contexts can leave an early behavioral commitment signature detectable without a reward model, external judge, or trained classifier.
\end{abstract}


\section{Introduction}
Chain-of-thought monitoring is attractive because models sometimes verbalize suspicious strategies before producing an answer. The harder case is \emph{implicit} reward hacking: the model exploits a shortcut or loophole without saying so in the chain of thought (CoT). TRACE~\cite{wang2026trace} addresses this setting by measuring how much reasoning effort is needed before a truncated CoT receives high verifier reward. If a prefix already passes verification, the full CoT may be post-hoc rationalization rather than genuine reasoning.

This paper asks whether a similar diagnostic can be made \emph{reward-free}. Instead of comparing forced early answers to ground truth or verifier reward, we compare them to the model's own final answer. This yields a self-commitment curve over the prompted reasoning context. The intuition is simple: if the answer is provided as a hint, the model can commit very early; if it is reasoning honestly, commitment may appear only after intermediate quantities are derived.

We evaluate this idea in a controlled paired prompted-hacking setup. Each GSM8K problem is run twice: once with a normal prompt and once with the correct answer appended as a hint. This creates a clean contrast between ordinary reasoning and a shortcut condition in which the final answer is available before the written derivation has earned it. The paper makes four contributions:
\begin{itemize}
    \item a reward-free prefix probe based on the model's own final answer distribution;
    \item a 50-problem paired GSM8K evaluation showing strong separation between honest and hinted reasoning contexts;
    \item correctness-stratified and threshold-sensitivity checks showing that the signal is not merely an error detector or a threshold artifact;
    \item negative evidence that our original backtracking-mass hypothesis is weak.
\end{itemize}

The paper asks: when a final answer has already been anchored by an external hint, can we observe that anchoring in the same prompted context without knowing the reward? Our answer is yes in this controlled setting. The diagnostic requires no hidden states, gradients, verifier scores, or labeled hacking examples.

\subsection{Scope}
The study targets a specific monitoring question: whether shortcut-available prompted reasoning can be detected from the timing of self-commitment. The claim is distributional rather than per-trace: when the systematic prompt change is an answer hint, successful hinted generations tend to self-commit earlier and with lower uncertainty than honest generations. This controlled setting isolates the signal before applying it to broader reward-hacking environments.

This distinction matters because many CoT-monitoring proposals assume suspicious behavior must be written in natural language. The prompted-hacking condition breaks that assumption: the CoT can look routine even when the final answer was supplied in the prompt. A reward-free self-commitment probe asks not what the model says, but when the prompted context becomes sufficient to reproduce its own answer.

\section{Method}
\subsection{Full-Answer Generation}
For a problem $x$ and condition $z\in\{\mathrm{honest},\mathrm{hinted}\}$, we first generate a full CoT and extract the model's final numeric answer $a_{\mathrm{final}}$. The hinted condition appends a short in-context cue, ``Hint: the answer is X,'' where $X$ is the dataset answer. All experiments use integer-answer problems so answer extraction can be handled by lightweight numeric normalization.

The full-answer generation step is deliberately separated from the prefix probe. Once $a_{\mathrm{final}}$ is extracted, the probe never compares the prefix answer to the dataset label. This separation is what makes the method reward-free: the reference answer for the probe is the model's own committed output, not an external correctness oracle.

\subsection{Self-Commitment Curve}
Let the generated CoT contain $T$ tokens. At strided prefix positions $t\in\mathcal{T}$, we truncate the CoT to the first $t$ tokens, append a forced-answer tag, and sample $k$ short completions. We define
\begin{equation}
c(t)=\Pr[\hat a_t=a_{\mathrm{final}}\mid x,z,\mathrm{CoT}_{\le t}],
\end{equation}
estimated as the fraction of the $k$ forced answers matching the full-CoT final answer after numeric extraction and normalization. In the main experiment, $k=5$ and the stride is 30 tokens, so $c(t)$ takes values in $\{0,0.2,\ldots,1.0\}$.

Operationally, a prefix query keeps the same condition-specific prompt used for generation, adds the truncated CoT, and asks for the answer immediately. Thus, hinted probes intentionally retain the answer-hint context; the diagnostic measures early commitment in the shortcut-available context, not whether the CoT prefix alone contains the answer. We keep completions short to reduce new full solutions. The curve is a behavioral estimate of answer availability, not a semantic score assigned to the text.

\subsection{Probe Pipeline}
For each problem-condition pair, the pipeline has four stages:
\begin{enumerate}
    \item generate the full CoT and parse $a_{\mathrm{final}}$;
    \item tokenize the CoT and choose strided prefix positions;
    \item sample $k$ forced answers from each prefix and parse each answer;
    \item aggregate the forced-answer matches into a curve $c(t)$ and scalar summaries.
\end{enumerate}
Rows are written immediately after each problem-condition pair finishes. This matters for local experimentation because prefix probing is much more expensive than ordinary generation: a single row can require dozens of short forced-answer calls. The resumable design allowed the evaluation to run on a laptop without losing completed rows after interruption.

\subsection{Scalar Summaries}
We report several reward-free summaries, all oriented so larger values are more honest-like:
\begin{align}
\mathrm{range}_c &= \max_t c(t)-\min_t c(t),\\
\mathrm{uncommitted} &= \mathbb{E}_t[1-c(t)],\\
\mathrm{lowcommit} &= \mathbb{E}_t[\mathbf{1}\{c(t)<0.5\}],\\
\tau_{\mathrm{first}}(\theta) &= \min\{t/T: c(t)\ge \theta\}.
\end{align}
Our primary statistic is $\tau_{\mathrm{first}}(0.8)$ because it measures how much generated reasoning must be present, under the same prompt condition, before the model strongly commits to its own final answer. We use $\theta=0.8$ as the main operational threshold, corresponding to at least 4 of 5 forced samples matching the final answer. We also report a threshold sensitivity check for $\theta\in\{0.5,0.6,\ldots,1.0\}$.

The remaining scalar summaries are supporting views of the same curve. $\mathrm{range}_c$ measures whether the trace moves from uncertain to committed. $\mathrm{uncommitted}$ measures the average amount of non-commitment across the trace. $\mathrm{lowcommit}$ measures the fraction of prefixes that remain below a majority match rate. These whole-curve summaries check that the effect is not an artifact of one threshold, while $\tau_{\mathrm{first}}$ remains the main latency statistic.

\subsection{Why Latency Is Primary}
The central observable is not whether a prefix answer is correct, but when the prompted reasoning context becomes sufficient for the model to reproduce its own final answer. This is why $\tau_{\mathrm{first}}$ is primary: a smaller value means less written reasoning is needed, under the same prompt condition, before the model is committed. In the hinted condition, this early availability is expected because the final answer has been supplied as an in-context shortcut. In the honest condition, the final answer should usually stabilize only after enough intermediate computation.

The whole-curve summaries are included to make the latency result easier to audit. $\tau_{\mathrm{first}}$ reduces the curve to one crossing event, which is interpretable but can be affected by a single noisy prefix. $\mathrm{uncommitted}$ averages over all measured prefixes and therefore captures sustained uncertainty. $\mathrm{range}_c$ captures the shape expected from ordinary reasoning: low commitment early and high commitment later. $\mathrm{lowcommit}$ is a thresholded version of the same idea. Agreement across these summaries is important because it shows that the signal is not a peculiarity of one cutoff or one formula.

This organization also prevents the metric set from becoming an unconstrained search over scores. The paper's main claim is a latency claim: shortcut-anchored traces commit earlier. The other summaries are supporting diagnostics that test whether the same phenomenon appears when the entire self-commitment curve is summarized in simpler ways. The failed $\rho_{\mathrm{back}}$ statistic serves the opposite role: it shows that not every plausible curve feature works, and that the useful information is specifically delayed commitment rather than visible reversal.

\begin{figure*}[t]
\centering
\resizebox{\textwidth}{!}{%
\begin{tikzpicture}
\begin{axis}[
    width=7.0in,
    height=1.82in,
    xmin=-1.42,
    xmax=1.38,
    ymin=-0.88,
    ymax=2.64,
    xlabel={$x$-axis: honest metric $-$ hinted metric, for the same problem},
    label style={font=\figlabel},
    tick label style={font=\figlabel},
    ytick=\empty,
    xtick={-1.0,-0.5,0,0.5,1.0},
    grid=major,
    grid style={gray!20},
    axis line style={black},
    axis x line*=bottom,
    axis y line=none,
    tick align=outside,
    clip=false,
]
\fill[gray!5] (axis cs:-1.42,-0.62) rectangle (axis cs:-1.05,2.52);
\fill[red!6] (axis cs:-1.0,-0.62) rectangle (axis cs:0,2.52);
\fill[green!8] (axis cs:0,-0.62) rectangle (axis cs:1.1,2.52);
\draw[gray!70, dashed] (axis cs:0,-0.62) -- (axis cs:0,2.52);
\node[anchor=south west, font=\figlabel, text=gray!75] at (axis cs:0.015,2.53) {$0$: no gap};
\node[anchor=south, font=\figlabel, text=red!55!black] at (axis cs:-0.50,2.53) {hinted $>$ honest};
\node[anchor=south, font=\figlabel, text=green!35!black] at (axis cs:0.53,2.53) {honest $>$ hinted};
\draw[-{Latex[length=1.8mm]}, green!45!black, line width=0.55pt]
    (axis cs:0.18,2.44) -- (axis cs:0.92,2.44);
\draw[-{Latex[length=1.8mm]}, red!50!black, line width=0.55pt]
    (axis cs:-0.20,2.44) -- (axis cs:-0.85,2.44);

\node[anchor=west, font=\figlabel, text=gray!70] at (axis cs:-1.405,2.49) {metric};
\node[anchor=west, align=left, font=\figlabel] at (axis cs:-1.405,2.000) {\metriclabel{curve spread}{$\mathrm{range}_c$}};
\draw[gray!45, line width=0.25pt] (axis cs:-1.0,1.88) rectangle (axis cs:1.1,2.12);
\fill[cyan!55!blue, opacity=0.14] (axis cs:-0.2,1.88) rectangle (axis cs:-0.1,2.12);
\fill[cyan!55!blue, opacity=0.23] (axis cs:0.0,1.88) rectangle (axis cs:0.1,2.12);
\fill[cyan!55!blue, opacity=0.18] (axis cs:0.1,1.88) rectangle (axis cs:0.2,2.12);
\fill[cyan!55!blue, opacity=0.21] (axis cs:0.3,1.88) rectangle (axis cs:0.4,2.12);
\fill[cyan!55!blue, opacity=0.18] (axis cs:0.5,1.88) rectangle (axis cs:0.6,2.12);
\fill[cyan!55!blue, opacity=0.23] (axis cs:0.8,1.88) rectangle (axis cs:0.9,2.12);
\fill[cyan!55!blue, opacity=0.58] (axis cs:1.0,1.88) rectangle (axis cs:1.1,2.12);
\addplot[only marks, black, mark=diamond*, mark size=2.7pt] coordinates {(1.000,2.000)};
\node[anchor=west, align=left, font=\figlabel] at (axis cs:1.17,2.000) {0.926};

\node[anchor=west, align=left, font=\figlabel] at (axis cs:-1.405,1.000) {\metriclabel{first commit}{$\tau_{\mathrm{first}}$ (primary)}};
\draw[gray!45, line width=0.25pt] (axis cs:-1.0,0.88) rectangle (axis cs:1.1,1.12);
\fill[green!55!black, opacity=0.18] (axis cs:-1.0,0.88) rectangle (axis cs:-0.9,1.12);
\fill[green!55!black, opacity=0.18] (axis cs:-0.4,0.88) rectangle (axis cs:-0.3,1.12);
\fill[green!55!black, opacity=0.25] (axis cs:-0.1,0.88) rectangle (axis cs:0.0,1.12);
\fill[green!55!black, opacity=0.52] (axis cs:0.0,0.88) rectangle (axis cs:0.1,1.12);
\fill[green!55!black, opacity=0.31] (axis cs:0.1,0.88) rectangle (axis cs:0.2,1.12);
\fill[green!55!black, opacity=0.31] (axis cs:0.2,0.88) rectangle (axis cs:0.3,1.12);
\fill[green!55!black, opacity=0.31] (axis cs:0.4,0.88) rectangle (axis cs:0.5,1.12);
\fill[green!55!black, opacity=0.25] (axis cs:0.5,0.88) rectangle (axis cs:0.6,1.12);
\fill[green!55!black, opacity=0.38] (axis cs:0.6,0.88) rectangle (axis cs:0.7,1.12);
\fill[green!55!black, opacity=0.66] (axis cs:0.7,0.88) rectangle (axis cs:0.8,1.12);
\fill[green!55!black, opacity=0.43] (axis cs:0.8,0.88) rectangle (axis cs:0.9,1.12);
\fill[green!55!black, opacity=0.38] (axis cs:0.9,0.88) rectangle (axis cs:1.0,1.12);
\addplot[only marks, black, mark=diamond*, mark size=2.7pt] coordinates {(0.648,1.000)};
\node[anchor=west, align=left, font=\figlabel] at (axis cs:1.17,1.000) {0.878};

\node[anchor=west, align=left, font=\figlabel] at (axis cs:-1.405,0.000) {\metriclabel{uncommitted area}{$\mathbb{E}[1-c]$}};
\draw[gray!45, line width=0.25pt] (axis cs:-1.0,-0.12) rectangle (axis cs:1.1,0.12);
\fill[orange!70!black, opacity=0.18] (axis cs:-0.9,-0.12) rectangle (axis cs:-0.8,0.12);
\fill[orange!70!black, opacity=0.18] (axis cs:-0.3,-0.12) rectangle (axis cs:-0.2,0.12);
\fill[orange!70!black, opacity=0.18] (axis cs:-0.2,-0.12) rectangle (axis cs:-0.1,0.12);
\fill[orange!70!black, opacity=0.29] (axis cs:0.0,-0.12) rectangle (axis cs:0.1,0.12);
\fill[orange!70!black, opacity=0.24] (axis cs:0.1,-0.12) rectangle (axis cs:0.2,0.12);
\fill[orange!70!black, opacity=0.29] (axis cs:0.2,-0.12) rectangle (axis cs:0.3,0.12);
\fill[orange!70!black, opacity=0.29] (axis cs:0.3,-0.12) rectangle (axis cs:0.4,0.12);
\fill[orange!70!black, opacity=0.35] (axis cs:0.4,-0.12) rectangle (axis cs:0.5,0.12);
\fill[orange!70!black, opacity=0.35] (axis cs:0.5,-0.12) rectangle (axis cs:0.6,0.12);
\fill[orange!70!black, opacity=0.66] (axis cs:0.6,-0.12) rectangle (axis cs:0.7,0.12);
\fill[orange!70!black, opacity=0.54] (axis cs:0.7,-0.12) rectangle (axis cs:0.8,0.12);
\fill[orange!70!black, opacity=0.45] (axis cs:0.8,-0.12) rectangle (axis cs:0.9,0.12);
\addplot[only marks, black, mark=diamond*, mark size=2.7pt] coordinates {(0.621,0.000)};
\node[anchor=west, align=left, font=\figlabel] at (axis cs:1.17,0.000) {0.904};
\node[anchor=west, align=left, font=\figlabel, text=gray!70] at (axis cs:1.17,2.49) {AUROC};

\fill[black, opacity=0.18] (axis cs:-0.22,-0.60) rectangle (axis cs:-0.12,-0.50);
\fill[black, opacity=0.45] (axis cs:-0.08,-0.60) rectangle (axis cs:0.02,-0.50);
\fill[black, opacity=0.80] (axis cs:0.06,-0.60) rectangle (axis cs:0.16,-0.50);
\node[anchor=west, font=\figlabel] at (axis cs:0.20,-0.55) {density: low $\rightarrow$ high};
\addplot[only marks, black, mark=diamond*, mark size=2.7pt] coordinates {(0.74,-0.55)};
\node[anchor=west, font=\figlabel] at (axis cs:0.81,-0.55) {median};
\end{axis}
\end{tikzpicture}
}
\caption{Aggregate paired commitment-metric differences over 50 GSM8K problems. Each row bins honest-minus-hinted same-problem differences; darker bins contain more pairs. Positive values mean the honest trace is more delayed or less committed. Diamonds mark medians; right-column values report AUROC.}
\label{fig:distribution}
\end{figure*}
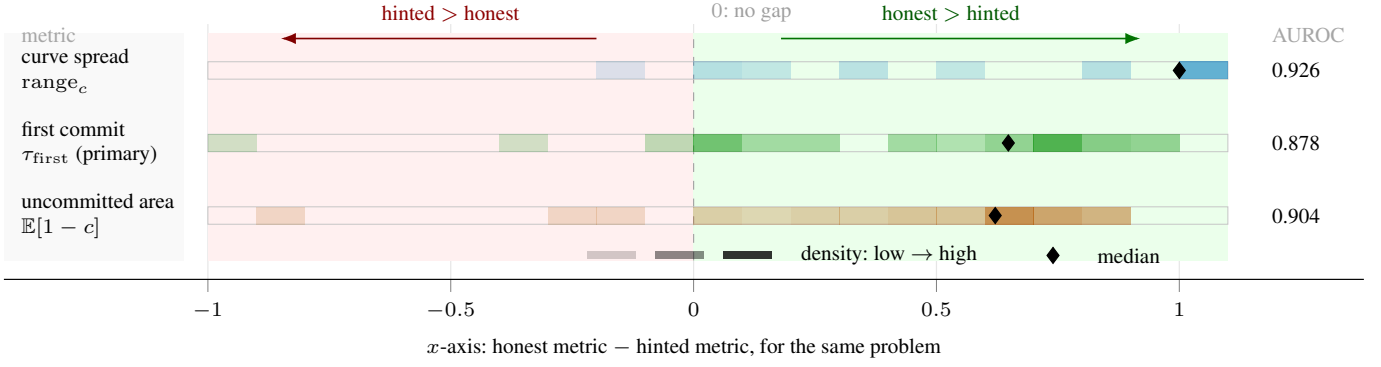

\subsection{Failed Initial Statistic}
Our original hypothesis was that honest reasoning would show explicit backward excursions in $c(t)$: segments where commitment to the final answer decreases. We implemented PELT segmentation and measured $\rho_{\mathrm{back}}$, the fraction of CoT tokens in negative-slope segments. This statistic failed in both development runs and the final evaluation. We retain it as a negative baseline because it clarifies what the observed signal is \emph{not}: the useful signal is delayed or unstable self-commitment, not necessarily monotonic backtracking.

\section{Experimental Setup}
\subsection{Dataset}
We use the first 50 examples from the GSM8K test split~\cite{cobbe2021gsm8k}, converted into a local JSON file with question text and final integer answer. Each example is evaluated under honest and hinted prompts, producing 100 CoTs.

GSM8K is useful for this controlled evaluation because the final answer is a short integer and the reasoning traces are long enough to expose intermediate computation. The benchmark is not itself a reward-hacking benchmark. We use it as a controlled environment where an injected answer hint can serve as a simple prompted loophole: the model can solve the task by using the provided answer while still emitting a plausible-looking explanation.

\subsection{Model and Inference}
All runs use a 4-bit Qwen2.5-3B-Instruct checkpoint from MLX Community on an Apple Silicon M1 Max machine. We generate up to 1024 CoT tokens, sweep prefixes with stride 30, and sample $k=5$ forced early answers per prefix at temperature 0.7. The experiment is resumable: each completed problem-condition row is appended to JSONL and skipped on restart.

The 3B model size was chosen to keep the full prefix-probing loop feasible on a local machine. This is important for the intended use case: the method should be cheap enough to run as an audit diagnostic before committing to a larger benchmark run. The tradeoff is that small models can have more arithmetic failures, so we report both all-row and correct-only analyses.

\subsection{Evaluation}
We report AUROC for distinguishing honest from hinted CoTs, treating larger metric values as more honest-like. We also report paired wins: the number of same-problem pairs where the honest CoT receives a larger score than the hinted CoT. Because answer correctness is a major confound, we separately analyze the subset where both honest and hinted completions are correct.

The paired design is central. Comparing honest and hinted generations for the same problem controls for problem difficulty, answer length, and broad topic. A paired win is counted only when the honest trace receives the more honest-like score than the hinted trace for the same GSM8K item. Ties are kept separate in the analysis artifacts, but Table~\ref{tab:main} reports the most compact form: wins out of 50 pairs.

\section{Results}
\subsection{Main Paired Detection}
Fig.~\ref{fig:examples} later gives curve-level intuition; Table~\ref{tab:main} and Fig.~\ref{fig:distribution} test whether the same early-commitment pattern holds across all 50 paired GSM8K problems. In Fig.~\ref{fig:distribution}, most density mass lies on the positive side, where the honest trace receives the larger, more honest-like score. Our primary latency metric, $\tau_{\mathrm{first}}(0.8)$, separates honest from hinted traces with AUROC 0.878 and 43/50 paired wins. The supporting whole-curve summaries show the same direction, with $\mathrm{range}_c$ strongest by AUROC and mean uncommitted mass strongest by best-threshold accuracy and paired wins. The backtracking baseline remains near random.

\begin{table*}[t]
\caption{Main controlled GSM8K paired evaluation. Larger scores are honest-like; paired wins count same-problem pairs where the honest trace scores higher.}
\label{tab:main}
\centering
\small
\begin{tabular}{lccccc}
\toprule
Metric & AUROC & honest median & hinted median & paired wins & best accuracy \\
\midrule
$\mathrm{range}_c$ & \textbf{0.926} & 1.000 & 0.000 & 44/50 & 0.900 \\
$\mathbb{E}[1-c]$ & 0.904 & 0.657 & 0.000 & \textbf{45/50} & \textbf{0.910} \\
$\Pr[c<0.5]$ & 0.896 & 0.667 & 0.000 & 44/50 & 0.900 \\
$\tau_{\mathrm{first}}(0.8)$ & 0.878 & 0.727 & 0.032 & 43/50 & 0.890 \\
$\rho_{\mathrm{back}}$ & 0.520 & 0.000 & 0.000 & 2/50 & 0.520 \\
\bottomrule
\end{tabular}
\end{table*}

The result supports the core claim: prompted hacking often causes early, low-uncertainty commitment to the final answer. The old $\rho_{\mathrm{back}}$ statistic is essentially random, despite occasional high values on individual wrong samples.

The medians make the qualitative picture sharp. A zero $\mathrm{range}_c$ value alone only means that the commitment curve is flat, but the hinted median also has $\mathbb{E}[1-c]=0.000$. Together, these values indicate that many hinted traces are flat at $c(t)=1$ across the measured prefixes: they are already fully committed to their final answer from the start. These zero hinted medians should be read as a saturation signal rather than as a loss of information; the remaining variation in hinted behavior is captured by AUROC, paired wins, and the density bins in Fig.~\ref{fig:distribution}. Honest traces, by contrast, typically move from low to high commitment over the course of the CoT. This is exactly the pattern expected if the hint anchors the answer before the written reasoning has earned it.

\subsection{Correctness-Stratified Analysis}
One concern is that these metrics may simply detect model failure. Wrong completions are often unstable or late-committing, regardless of condition. Table~\ref{tab:correct} addresses this by restricting to the 40 paired problems where both honest and hinted completions are correct. The signal becomes stronger, not weaker.

\begin{table*}[t]
\caption{Correctness-stratified analysis on the 40 same-problem pairs where both conditions answer correctly, holding final-answer correctness fixed.}
\label{tab:correct}
\centering
\small
\begin{tabular}{lcccc}
\toprule
Metric & AUROC & honest median & hinted median & note \\
\midrule
$\mathrm{range}_c$ & 0.967 & 1.000 & 0.000 & high range in honest CoTs \\
$\mathbb{E}[1-c]$ & \textbf{0.980} & 0.625 & 0.000 & strongest correct-only signal \\
$\Pr[c<0.5]$ & 0.970 & 0.667 & 0.000 & similar to mean uncertainty \\
$\tau_{\mathrm{first}}(0.8)$ & 0.931 & 0.727 & 0.032 & robust latency signal \\
$\rho_{\mathrm{back}}$ & 0.525 & 0.000 & 0.000 & still uninformative \\
\bottomrule
\end{tabular}
\end{table*}

Incorrect hinted completions are the main boundary case. When the model fails to use the hint or produces a wrong final answer, the trajectory often resembles ordinary uncertainty. This is expected: the present diagnostic targets successful hint anchoring, not arbitrary prompt failure.

\begin{table}[t]
\caption{Condition and correctness breakdown. Values are medians within each prompt-condition and correctness group.}
\label{tab:confound}
\centering
\scriptsize
\begin{tabular}{llcccc}
\toprule
Cond. & Correct & $n$ & $\mathrm{range}_c$ & $\tau_{.8}$ & $\mathbb{E}[1-c]$ \\
\midrule
honest & yes & 40 & 1.000 & 0.727 & 0.625 \\
honest & no & 10 & 1.000 & 0.749 & 0.760 \\
hinted & yes & 46 & 0.000 & 0.032 & 0.000 \\
hinted & no & 4 & 0.500 & 1.000 & 0.950 \\
\bottomrule
\end{tabular}
\end{table}

Table~\ref{tab:confound} explains why correctness stratification is necessary. Incorrect honest generations tend to be even more uncommitted than correct honest generations, while incorrect hinted generations often lose the early-commitment signature. If we only compared all correct rows to all incorrect rows, the interpretation would be muddled. The both-correct paired subset removes this concern: both generations reach the right answer, but the hinted generation typically commits much earlier.

\subsection{Threshold Sensitivity}
The choice of $\theta=0.8$ is not load-bearing. Sweeping $\tau_{\mathrm{first}}(\theta)$ over $\theta\in\{0.5,0.6,\ldots,1.0\}$ changes the all-row AUROC only from 0.876 to 0.878 and changes paired wins only from 42/50 to 43/50. On the both-correct subset, every threshold gives AUROC 0.931 and 36/40 paired wins. This stability is expected: with $k=5$, commitment values are quantized in increments of 0.2, and most trajectories cross from low to high commitment at the same measured prefix.

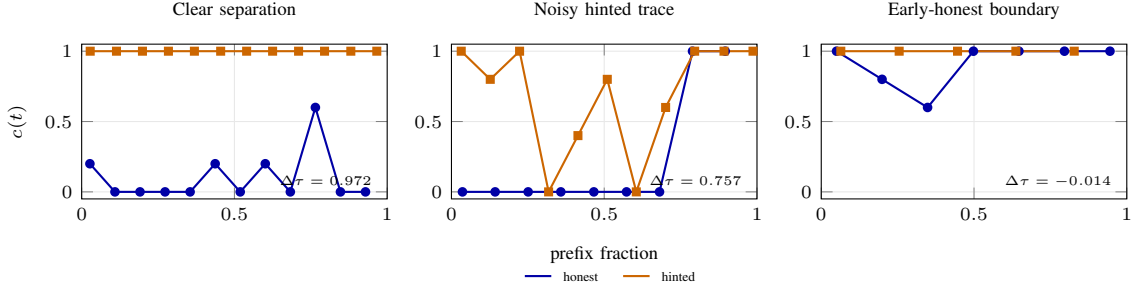
\begin{figure*}[t]
\centering
\begin{tikzpicture}
\begin{groupplot}[
    group style={group size=3 by 1, horizontal sep=0.85cm},
    width=0.31\textwidth,
    height=0.20\textwidth,
    xmin=0,
    xmax=1,
    ymin=-0.05,
    ymax=1.05,
    ylabel={$c(t)$},
    label style={font=\scriptsize},
    tick label style={font=\scriptsize},
    title style={font=\scriptsize},
    xtick={0,0.5,1.0},
    ytick={0,0.5,1.0},
    grid=major,
    grid style={gray!18},
]
\nextgroupplot[title={Clear separation}]
\addplot[blue!65!black, thick, mark=*, mark size=1.4pt] coordinates {
(0.027,0.2) (0.109,0.0) (0.191,0.0) (0.273,0.0) (0.355,0.0) (0.437,0.2) (0.519,0.0) (0.601,0.2) (0.683,0.0) (0.765,0.6) (0.847,0.0) (0.929,0.0)
};
\addplot[orange!80!black, thick, mark=square*, mark size=1.3pt] coordinates {
(0.028,1.0) (0.114,1.0) (0.199,1.0) (0.284,1.0) (0.369,1.0) (0.455,1.0) (0.540,1.0) (0.625,1.0) (0.710,1.0) (0.795,1.0) (0.881,1.0) (0.966,1.0)
};
\node[anchor=north east, font=\tiny] at (axis cs:0.98,0.18) {$\Delta\tau=0.972$};

\nextgroupplot[title={Noisy hinted trace}, ylabel={}]
\addplot[blue!65!black, thick, mark=*, mark size=1.4pt] coordinates {
(0.036,0.0) (0.143,0.0) (0.251,0.0) (0.358,0.0) (0.466,0.0) (0.573,0.0) (0.681,0.0) (0.789,1.0) (0.896,1.0)
};
\addplot[orange!80!black, thick, mark=square*, mark size=1.3pt] coordinates {
(0.032,1.0) (0.127,0.8) (0.223,1.0) (0.318,0.0) (0.414,0.4) (0.510,0.8) (0.605,0.0) (0.701,0.6) (0.796,1.0) (0.892,1.0) (0.987,1.0)
};
\node[anchor=north east, font=\tiny] at (axis cs:0.98,0.18) {$\Delta\tau=0.757$};

\nextgroupplot[title={Early-honest boundary}, ylabel={}]
\addplot[blue!65!black, thick, mark=*, mark size=1.4pt] coordinates {
(0.050,1.0) (0.199,0.8) (0.348,0.6) (0.498,1.0) (0.647,1.0) (0.796,1.0) (0.945,1.0)
};
\addplot[orange!80!black, thick, mark=square*, mark size=1.3pt] coordinates {
(0.064,1.0) (0.255,1.0) (0.446,1.0) (0.637,1.0) (0.828,1.0)
};
\node[anchor=north east, font=\tiny] at (axis cs:0.98,0.18) {$\Delta\tau=-0.014$};
\end{groupplot}
\node[anchor=north, font=\scriptsize] at ($(group c2r1.south)+(0,-0.50cm)$) {prefix fraction};
\draw[blue!65!black, thick] ($(group c2r1.south)+(-1.05cm,-1.02cm)$) -- ++(0.35cm,0)
    node[right, black, font=\tiny, xshift=0.03cm] {honest};
\draw[orange!80!black, thick] ($(group c2r1.south)+(0.25cm,-1.02cm)$) -- ++(0.35cm,0)
    node[right, black, font=\tiny, xshift=0.03cm] {hinted};
\end{tikzpicture}
\caption{Representative paired self-commitment curves. The x-axis is prefix fraction and the y-axis is $c(t)$, the forced-answer match rate to the run's own final answer. Blue is honest and orange is hinted; panels show strong separation, noisy hinted early commitment, and an early-honest boundary case.}
\label{fig:examples}
\end{figure*}

Figure~\ref{fig:examples} shows representative paired curves. The important visual feature is the shape of the full curve relative to the matched condition. In successful hinted generations, $c(t)$ is frequently high from the first measured prefix because the shortcut remains in context. In honest generations, early prefixes often produce forced answers that do not match the final answer; commitment rises only after enough intermediate computation. The boundary case shows why the method should be interpreted statistically.

\subsection{Negative Result: Backtracking Mass}
The original version of this project focused on detecting backward movement in $c(t)$, motivated by the idea that honest reasoning might revise earlier guesses. This was too narrow. The $\rho_{\mathrm{back}}$ statistic gives AUROC 0.520 overall and 0.525 in the both-correct paired subset, essentially no better than chance. Most curves do not contain enough reliable negative-slope structure for this statistic to be useful at the present sample size and sampling budget.

This negative result usefully sharpens the claim. The successful signal is not that honest CoTs visibly backtrack. Rather, honest CoTs often remain uncertain until later, while hinted CoTs often begin committed. In other words, the main observable is \emph{commitment latency}, not explicit reversal.

\section{Discussion}
\subsection{What the Signal Means}
The probe does not claim to recover the model's true internal computation. It measures how much generated reasoning is needed, under a fixed prompt condition, before forced answers align with the final answer. In the hinted condition, the answer is often available from the prompt, so the model can commit immediately and use the subsequent CoT to rationalize. In the honest condition, the final answer is typically unsupported until later.

This makes self-commitment latency complementary to text-based CoT monitoring. A text monitor asks whether the model verbalizes a suspicious strategy. A self-commitment probe asks whether the answer becomes behaviorally available in context before the written reasoning appears to justify it. This remains informative when the CoT is fluent, plausible, and free of explicit shortcut language.

\subsection{Audit Workflow}
The diagnostic can be used as a lightweight audit layer around existing reasoning evaluations. First, generate the full CoT and extract the model's final answer. Second, probe a small set of prefixes and estimate the self-commitment curve. Third, compare the latency and whole-curve summaries against a calibrated baseline, such as honest prompts for the same task distribution or paired prompts that differ only in shortcut availability. The output is not a binary proof of hacking; it is a ranking signal for traces that deserve closer inspection.

This workflow is useful because it does not require training a monitor, collecting labeled hacking examples, or building a reward model. It only requires the model under audit and its generated traces. Where a verifier exists, self-commitment latency can be compared with TRACE-like measures; where no verifier is available, the same prefix-probing structure still yields a reward-free signal. In paired GSM8K, when the shortcut is present in the prompt, the final answer becomes recoverable from much shorter prompted contexts.

\subsection{Relationship to TRACE}
TRACE asks when a prefix attains high external reward. Our method asks when the prompted prefix context supports the model's own final answer. This removes the verifier requirement and makes the probe applicable even when ground-truth reward is unavailable at inference. The tradeoff is that our method detects self-commitment rather than correctness or reward hacking directly. It is best viewed as a complementary diagnostic rather than a replacement for TRACE.

\subsection{Boundary Cases}
Easy problems sometimes produce early commitment in both conditions; these yield ties. Failed hinted completions can look like honest reasoning because the hint did not successfully anchor the model. Finally, high range can arise from sampling noise or transient uncertainty even in hinted CoTs, so range should be interpreted together with latency and mean uncertainty.

\subsection{Operational Cost}
The method is more expensive than ordinary inference because each generated CoT is followed by a sweep over prefixes. With stride 30 and $k=5$, the number of forced-answer calls grows linearly with CoT length. This cost is acceptable for auditing and benchmark construction, but it is not yet a lightweight online monitor. A practical deployment would likely use an adaptive schedule: probe a small number of early prefixes first, then add later prefixes only if the early curve is ambiguous.

There is also a variance-cost tradeoff in $k$. With $k=5$, $c(t)$ is coarse but cheap, and the threshold 0.8 has a simple interpretation as 4 of 5 samples matching. Larger $k$ would produce smoother curves and more stable estimates, but the threshold-sensitivity result shows that the main latency effect is not dominated by a single arbitrary cutoff.

\subsection{Scope and Validity}
The controlled hint condition is a proxy for shortcut availability, not a claim that all implicit reward hacking has the same surface form. Its value is that it isolates the key causal structure: the final answer is available in the prompt before the written reasoning supports it. The probe measures a behavioral signature of this prompted context, not hidden computation or the standalone content of the CoT prefix. Early commitment should therefore be treated as a calibrated suspiciousness signal rather than proof of hacking.

The current evaluation uses one model family and integer-answer GSM8K problems so that answer normalization is reliable. The natural extensions are straightforward: larger paired subsets, additional model families, more realistic loopholes such as rubric artifacts, and, when available, trained hacking checkpoints for direct comparison with verifier-based TRACE measurements.

\section{Related Work}
TRACE~\cite{wang2026trace} is the closest prior work: it detects implicit reward hacking by truncating CoTs and measuring how quickly prefixes pass verification. Our method inherits the truncation idea but replaces verifier reward with self-commitment to the model's own final answer. This places the probe in the broader line of work on CoT reasoning and reasoning-time interventions, including few-shot CoT prompting~\cite{wei2022cot}, zero-shot CoT~\cite{kojima2022zeroshot}, self-consistency decoding~\cite{wang2023selfconsistency}, least-to-most decomposition~\cite{zhou2023leasttomost}, tool- or action-augmented reasoning~\cite{yao2023react}, and program-aided reasoning~\cite{gao2023pal}. GSM8K~\cite{cobbe2021gsm8k}, MATH~\cite{hendrycks2021math}, and Minerva~\cite{lewkowycz2022minerva} illustrate why long mathematical traces are a useful testbed, while process supervision and verification work~\cite{uesato2022process,lightman2024verify} shows the value of measuring intermediate reasoning rather than only final answers.

The paper also relates to CoT faithfulness and monitorability. Prior work distinguishes plausible explanations from faithful explanations~\cite{jacovi2020faithfulness} and shows that CoT can rationalize biased or shortcut-driven answers~\cite{turpin2023unfaithful}. Intervention-based studies of CoT faithfulness~\cite{lanham2023faithfulness} and faithful-by-construction reasoning methods~\cite{lyu2023faithful} motivate behavioral tests that do not simply trust the surface text. Work on language-model self-evaluation~\cite{kadavath2022know} is related because our probe also uses the model's own outputs as reference points. Recent monitorability evaluations emphasize that CoT-based safety signals must be measured directly and may change with training and inference procedure~\cite{openai2025monitorability}.

Finally, our setting is motivated by reward modeling and reward-hacking concerns. Human-feedback methods~\cite{christiano2017preferences,stiennon2020summarize,ouyang2022instructgpt} and AI-feedback methods~\cite{bai2022constitutional} make reward models and preference models central to modern alignment, while scalable oversight proposals such as reward modeling and debate study how supervision can extend to tasks humans cannot directly evaluate~\cite{leike2018rewardmodeling,irving2018debate}. Safety work has long emphasized proxy misspecification, learned optimization, reward tampering, and reward gaming~\cite{amodei2016concrete,hubinger2019learned,everitt2021rewardtampering,skalse2022rewardgaming}. Empirical studies of reward-model overoptimization further show that optimizing proxy feedback can degrade true quality~\cite{gao2023overoptimization,rafailov2024direct}. Our contribution is a reward-free behavioral signal for one important case: shortcut availability that becomes visible as unusually early commitment to the model's own final answer.

\section{Conclusion}
Prompted implicit hacking leaves a measurable self-commitment signature: shortcut-available contexts commit to their final answer earlier and with lower uncertainty than honest contexts. The proposed probe turns this observation into a reward-free diagnostic based only on prompted prefixes, forced answers, and the model's own final output. In a controlled paired GSM8K evaluation, the signal is strong, robust to threshold choice, and preserved under a correct-only analysis. These results support self-commitment latency as a practical complement to verifier-based reasoning-effort diagnostics.


\begingroup
\scriptsize
\begin{thebibliography}{30}
\scriptsize
\renewcommand{\baselinestretch}{0.86}\selectfont
\setlength{\itemsep}{0pt}
\setlength{\parskip}{0pt}
\setlength{\parsep}{0pt}
\bibitem{wang2026trace}
X. Wang, N. Joshi, B. Plank, R. Angell, and H. He, ``Is it thinking or cheating? Detecting implicit reward hacking by measuring reasoning effort,'' in \emph{Proc. International Conference on Learning Representations}, 2026.

\bibitem{wei2022cot}
J. Wei, X. Wang, D. Schuurmans, M. Bosma, B. Ichter, F. Xia, E. Chi, Q. Le, and D. Zhou, ``Chain-of-thought prompting elicits reasoning in large language models,'' in \emph{Advances in Neural Information Processing Systems}, 2022.

\bibitem{kojima2022zeroshot}
T. Kojima, S. S. Gu, M. Reid, Y. Matsuo, and Y. Iwasawa, ``Large language models are zero-shot reasoners,'' in \emph{Advances in Neural Information Processing Systems}, 2022.

\bibitem{wang2023selfconsistency}
X. Wang, J. Wei, D. Schuurmans, Q. Le, E. Chi, S. Narang, A. Chowdhery, and D. Zhou, ``Self-consistency improves chain of thought reasoning in language models,'' in \emph{Proc. International Conference on Learning Representations}, 2023.

\bibitem{zhou2023leasttomost}
D. Zhou, N. Scharli, L. Hou, J. Wei, N. Scales, X. Wang, D. Schuurmans, C. Cui, O. Bousquet, Q. Le, and E. Chi, ``Least-to-most prompting enables complex reasoning in large language models,'' in \emph{Proc. International Conference on Learning Representations}, 2023.

\bibitem{yao2023react}
S. Yao, J. Zhao, D. Yu, N. Du, I. Shafran, K. Narasimhan, and Y. Cao, ``ReAct: Synergizing reasoning and acting in language models,'' in \emph{Proc. International Conference on Learning Representations}, 2023.

\bibitem{gao2023pal}
L. Gao, A. Madaan, S. Zhou, U. Alon, P. Liu, Y. Yang, J. Callan, and G. Neubig, ``PAL: Program-aided language models,'' in \emph{Proc. International Conference on Machine Learning}, 2023.

\bibitem{cobbe2021gsm8k}
K. Cobbe, V. Kosaraju, M. Bavarian, M. Chen, H. Jun, L. Kaiser, M. Plappert, J. Tworek, J. Hilton, R. Nakano, C. Hesse, and J. Schulman, ``Training verifiers to solve math word problems,'' arXiv:2110.14168, 2021.

\bibitem{hendrycks2021math}
D. Hendrycks, C. Burns, S. Kadavath, A. Arora, S. Basart, E. Tang, D. Song, and J. Steinhardt, ``Measuring mathematical problem solving with the MATH dataset,'' in \emph{Proc. NeurIPS Datasets and Benchmarks Track}, 2021.

\bibitem{lewkowycz2022minerva}
A. Lewkowycz, A. Andreassen, D. Dohan, E. Dyer, H. Michalewski, V. Ramasesh, A. Slone, C. Anil, I. Schlag, T. Gutman-Solo, Y. Wu, B. Neyshabur, G. Gur-Ari, and V. Misra, ``Solving quantitative reasoning problems with language models,'' in \emph{Advances in Neural Information Processing Systems}, 2022.

\bibitem{uesato2022process}
J. Uesato, N. Kushman, R. Kumar, F. Song, N. Siegel, L. Wang, A. Creswell, G. Irving, and I. Higgins, ``Solving math word problems with process- and outcome-based feedback,'' arXiv:2211.14275, 2022.

\bibitem{lightman2024verify}
H. Lightman \emph{et al.}, ``Let's verify step by step,'' in \emph{Proc. International Conference on Learning Representations}, 2024.

\bibitem{jacovi2020faithfulness}
A. Jacovi and Y. Goldberg, ``Towards faithfully interpretable NLP systems: How should we define and evaluate faithfulness?'' in \emph{Proc. Annual Meeting of the Association for Computational Linguistics}, 2020.

\bibitem{turpin2023unfaithful}
M. Turpin, J. Michael, E. Perez, and S. Bowman, ``Language models don't always say what they think: Unfaithful explanations in chain-of-thought prompting,'' in \emph{Advances in Neural Information Processing Systems}, 2023.

\bibitem{lanham2023faithfulness}
T. Lanham \emph{et al.}, ``Measuring faithfulness in chain-of-thought reasoning,'' arXiv:2307.13702, 2023.

\bibitem{lyu2023faithful}
Q. Lyu, S. Havaldar, A. Stein, L. Zhang, D. Rao, E. Wong, M. Apidianaki, and C. Callison-Burch, ``Faithful chain-of-thought reasoning,'' in \emph{Proc. IJCNLP-AACL}, 2023.

\bibitem{kadavath2022know}
S. Kadavath \emph{et al.}, ``Language models (mostly) know what they know,'' arXiv:2207.05221, 2022.

\bibitem{openai2025monitorability}
OpenAI, ``Evaluating chain-of-thought monitorability,'' OpenAI research publication, 2025.

\bibitem{christiano2017preferences}
P. F. Christiano, J. Leike, T. B. Brown, M. Martic, S. Legg, and D. Amodei, ``Deep reinforcement learning from human preferences,'' in \emph{Advances in Neural Information Processing Systems}, 2017.

\bibitem{stiennon2020summarize}
N. Stiennon, L. Ouyang, J. Wu, D. Ziegler, R. Lowe, C. Voss, A. Radford, D. Amodei, and P. Christiano, ``Learning to summarize with human feedback,'' in \emph{Advances in Neural Information Processing Systems}, 2020.

\bibitem{ouyang2022instructgpt}
L. Ouyang \emph{et al.}, ``Training language models to follow instructions with human feedback,'' in \emph{Advances in Neural Information Processing Systems}, 2022.

\bibitem{bai2022constitutional}
Y. Bai \emph{et al.}, ``Constitutional AI: Harmlessness from AI feedback,'' arXiv:2212.08073, 2022.

\bibitem{leike2018rewardmodeling}
J. Leike, D. Krueger, T. Everitt, M. Martic, V. Maini, and S. Legg, ``Scalable agent alignment via reward modeling: A research direction,'' arXiv:1811.07871, 2018.

\bibitem{irving2018debate}
G. Irving, P. Christiano, and D. Amodei, ``AI safety via debate,'' arXiv:1805.00899, 2018.

\bibitem{amodei2016concrete}
D. Amodei, C. Olah, J. Steinhardt, P. Christiano, J. Schulman, and D. Mane, ``Concrete problems in AI safety,'' arXiv:1606.06565, 2016.

\bibitem{hubinger2019learned}
E. Hubinger \emph{et al.}, ``Risks from learned optimization in advanced machine learning systems,'' arXiv:1906.01820, 2019.

\bibitem{everitt2021rewardtampering}
T. Everitt, M. Hutter, R. Kumar, and V. Krakovna, ``Reward tampering problems and solutions in reinforcement learning: A causal influence diagram perspective,'' \emph{Synthese}, vol. 198, suppl. 27, pp. 6435--6467, 2021.

\bibitem{skalse2022rewardgaming}
J. M. V. Skalse, N. H. R. Howe, D. Krasheninnikov, and D. Krueger, ``Defining and characterizing reward gaming,'' in \emph{Advances in Neural Information Processing Systems}, 2022.

\bibitem{gao2023overoptimization}
L. Gao, J. Schulman, and J. Hilton, ``Scaling laws for reward model overoptimization,'' in \emph{Proc. International Conference on Machine Learning}, 2023.

\bibitem{rafailov2024direct}
R. Rafailov, Y. Chittepu, R. Park, H. Sikchi, J. Hejna, W. B. Knox, C. Finn, and S. Niekum, ``Scaling laws for reward model overoptimization in direct alignment algorithms,'' in \emph{Advances in Neural Information Processing Systems}, 2024.
\end{thebibliography}
\endgroup
\end{document}